\documentclass[runningheads]{llncs}
\usepackage[T1]{fontenc}
\usepackage{graphicx}
\usepackage{booktabs}

\usepackage[table]{xcolor}
\usepackage{pifont}
\newcommand{\cmark}{\textcolor{green}{\ding{51}}}
\newcommand{\xmark}{\textcolor{red}{\ding{55}}}

\usepackage{background}

\newcommand{\mywatermark}{%
    \begin{minipage}{\textwidth}
        \centering
        \fontsize{10}{10}\selectfont 
        This paper has been accepted at the 10th International Conference on Computer Vision \& Image Processing (CVIP 2025)
    \end{minipage}%
}

\backgroundsetup{
    scale=1,
    color=gray,
    angle=0,
    opacity=0.5,
    position=current page.north,
    vshift=-1cm, 
    contents={\mywatermark}
}

\begin{document}
\title{How Real is CARLA’s Dynamic Vision Sensor? \\ A  Study on the Sim-to-Real Gap in Traffic Object Detection}
\titlerunning{How Real is CARLA’s Dynamic Vision Sensor?}
%
\author{Kaiyuan Tan\inst{1}\orcidID{0009-0004-8640-0036} \and
Pavan Kumar B N\inst{2}\orcidID{0000-0002-5204-8036} \and
Bharatesh Chakravarthi\inst{1}\orcidID{0000-0002-4978-434X}}
\authorrunning{Tan. et al.}
%
\institute{Arizona State University, Tempe, AZ 85281, USA \\ \email{ktan24@asu.edu, bharatesh@asu.edu}\and
Indian Institute of Information Technology, Sri City, Chittoor, AP, India
\email{pavanbn8@gmail.com}\\
 }
\maketitle              
\begin{abstract}
Event cameras are gaining traction in traffic monitoring applications due to their low latency, high temporal resolution, and energy efficiency, which makes them well-suited for real-time object detection at traffic intersections. However, the development of robust event-based detection models is hindered by the limited availability of annotated real-world datasets. To address this, several simulation tools have been developed to generate synthetic event data. Among these, the CARLA driving simulator includes a built-in dynamic vision sensor (DVS) module that emulates event camera output. Despite its potential, the sim-to-real gap for event-based object detection remains insufficiently studied.
In this work, we present a systematic evaluation of this gap by training a recurrent vision transformer model exclusively on synthetic data generated using CARLA’s DVS and testing it on varying combinations of synthetic and real-world event streams. Our experiments show that models trained solely on synthetic data perform well on synthetic-heavy test sets but suffer significant performance degradation as the proportion of real-world data increases. In contrast, models trained on real-world data demonstrate stronger generalization across domains.
This study offers the first quantifiable analysis of the sim-to-real gap in event-based object detection using CARLA’s DVS. Our findings highlight limitations in current DVS simulation fidelity and underscore the need for improved domain adaptation techniques in neuromorphic vision for traffic monitoring. 

\end{abstract}

\section{Introduction}
Event-based cameras~\cite{chakravarthi2024recent} are being actively explored in traffic monitoring applications due to their unique ability to capture fast-moving objects with low latency, high temporal resolution, and energy efficiency~\cite{shariff2024event}. Unlike conventional frame-based cameras that capture scenes at fixed intervals, event cameras asynchronously record changes in pixel intensity, producing a continuous stream of events, making them well-suited for real-time object detection in complex, high-speed environments such as urban intersections, where timely and accurate perception is critical for tasks like vehicle tracking, pedestrian safety, and adaptive traffic management~\cite{gallego2020event,chakravarthi2023event}.
Despite these advantages, the development of robust event-based object detection models for traffic scenarios is significantly constrained by the scarcity of annotated real-world event datasets. Labeling event data is particularly challenging, often requiring synchronized recordings from both event and frame-based cameras and subsequent projection of annotations, an effort-intensive and error-prone process. As a result, large-scale training and evaluation of machine learning models for event-based perception remain limited in practice.

\begin{figure}[t]
\centering
\includegraphics[width=0.95\linewidth]{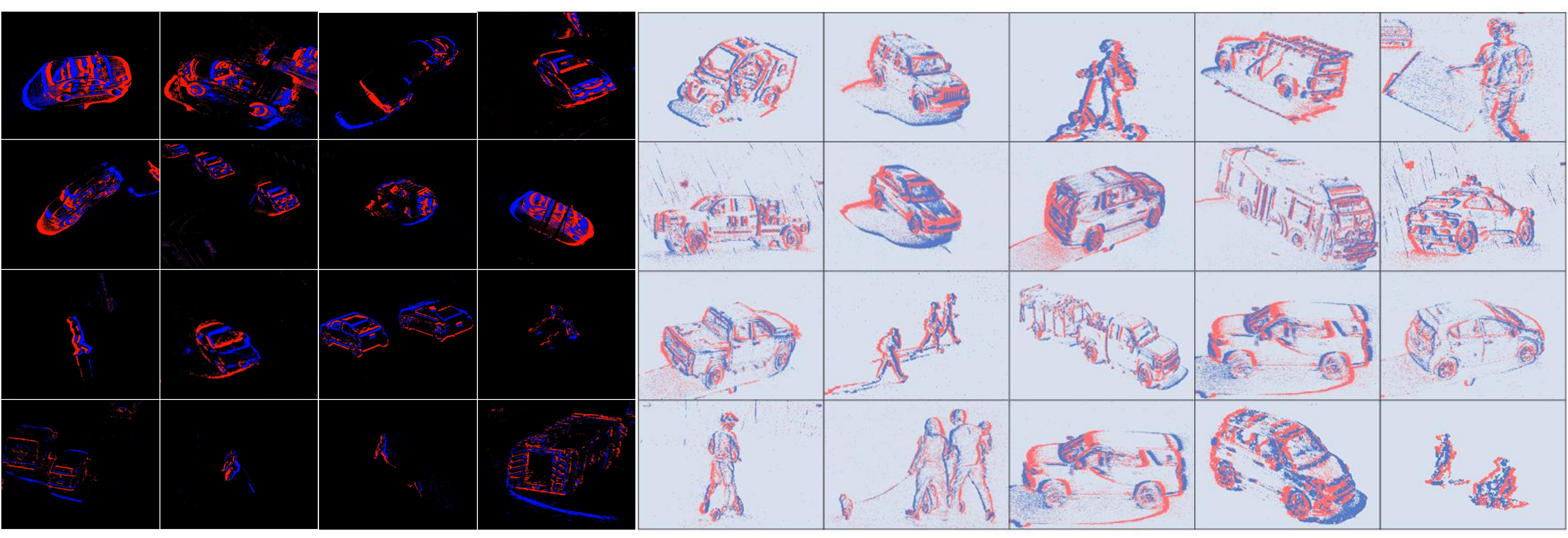}
\caption{Examples of traffic objects captured using event-based vision. (Left, black background) Synthetic events generated by CARLA’s DVS module. (Right, gray background) Real-world events from the eTram dataset~\cite{verma2024etram}, recorded using a physical event camera (Prophesee's EVK $4$ HD).}
\label{fig01}
\end{figure}

To mitigate this challenge, researchers have proposed a number of event camera simulators capable of generating synthetic event data~\cite{bi2017pix2nvs,dosovitskiy2017carla,han2024physical,hu2021v2e,joubert2021event,lin2022dvs,mueggler2017event}. 
Among them, CARLA~\cite{dosovitskiy2017carla} stands out as a widely used open-source driving simulator that provides a high-fidelity, controllable environment for traffic scenario modeling. Importantly, CARLA includes a built-in dynamic vision sensor (DVS) module that emulates the output of real event cameras. This makes it uniquely suited for our study, as it not only supports complex, multi-agent traffic interactions but also enables event-based data collection in diverse lighting, weather, and traffic conditions. \textit{CARLA was chosen for this work because it offers a rare combination of realistic traffic simulation and native support for event camera emulation, making it an ideal platform for evaluating synthetic data in traffic monitoring contexts}~\cite{aliminati2024sevd}. Fig.~\ref{fig01} shows examples of traffic objects captured using synthetic event data from CARLA and real-world data from the eTram~\cite{verma2024etram} dataset, both of which are used in this study to systematically assess the sim-to-real gap in event-based object detection.

In this study, we evaluate the realism and reliability of CARLA’s DVS-generated event data for object detection tasks in traffic environments. Specifically, we train a recurrent vision transformer (RVT)~\cite{gehrig2023recurrent}, a state-of-the-art model tailored for event-based vision, exclusively on synthetic data produced by CARLA and evaluate its performance on varying combinations of synthetic and real-world test data. \textit{The RVT was selected for its strong performance on sparse spatiotemporal data and its architectural suitability for handling asynchronous event streams, enabling a rigorous test of the data’s generalizability.}
Through this sim-to-real evaluation, we aim to determine whether CARLA’s synthetic event data can serve as an effective proxy for real-world conditions or if substantial performance discrepancies persist. To our knowledge, this is the first study to provide a  quantification of the sim-to-real gap using CARLA’s native DVS module in a traffic object detection setting.
We summarize the key contributions of the paper below:

\begin{itemize}

\item \textit{Synthetic Event-Based Vision Dataset for Traffic Monitoring:} We offer a dataset captured from a fixed perception setup at multiple intersection scenarios. It includes diverse conditions spanning various weather patterns, lighting environments, and traffic densities, along with detailed annotations of traffic objects, enabling realistic benchmarking and training for event-based perception models.
        
\item \textit{Evaluation of CARLA’s DVS for Synthetic Data Generation:} We assess the quality and applicability of CARLA’s synthetic event data in the context of traffic monitoring, specifically focusing on object detection performance.

\item \textit{Sim-to-Real Gap Analysis and Benchmarking:} We quantify the domain gap between synthetic and real-world event data by evaluating model performance across both domains. Our findings highlight the limitations of current DVS simulation fidelity in CARLA and establish a baseline for future research on domain adaptation and transfer learning in neuromorphic vision for traffic monitoring.

\end{itemize}

\section{Related Study}
This section reviews the foundational principles of event cameras, event camera simulators developed to mitigate the scarcity of annotated real-world data, and outlines the rationale for selecting CARLA’s DVS as a synthetic data source for traffic object detection in this study.

\subsection{Event Cameras}
Event cameras are bio-inspired vision sensors that detect per-pixel changes in brightness asynchronously. Unlike conventional frame-based cameras that capture full images at fixed intervals, event cameras emit a continuous stream of events whenever the logarithmic intensity change at a pixel exceeds a threshold. Each event is encoded as a tuple $\langle x, y, t, p \rangle$, representing the pixel coordinates $(x, y)$, timestamp $t$, and polarity $p$, which indicates whether the intensity increased or decreased \cite{chakravarthi2024recent,gallego2020event}.

This sensing paradigm offers several advantages over conventional RGB cameras, including high dynamic range, negligible motion blur, microsecond-level temporal resolution, and extremely low power consumption. These characteristics make event cameras particularly well-suited for real-time perception in high-speed, dynamic, and low-light environments. As a result, their adoption is expanding across a wide range of domains, including robotics, autonomous driving, augmented and virtual reality, mobile and wearable devices, Internet of Things, medical imaging, positioning and navigation systems, $3D$ scanning and surface profiling, as well as defense and surveillance applications~\cite{prophesee2024applications,inivation_solutions}.

\begin{figure}[t]
\centering
\includegraphics[width=0.70\linewidth]{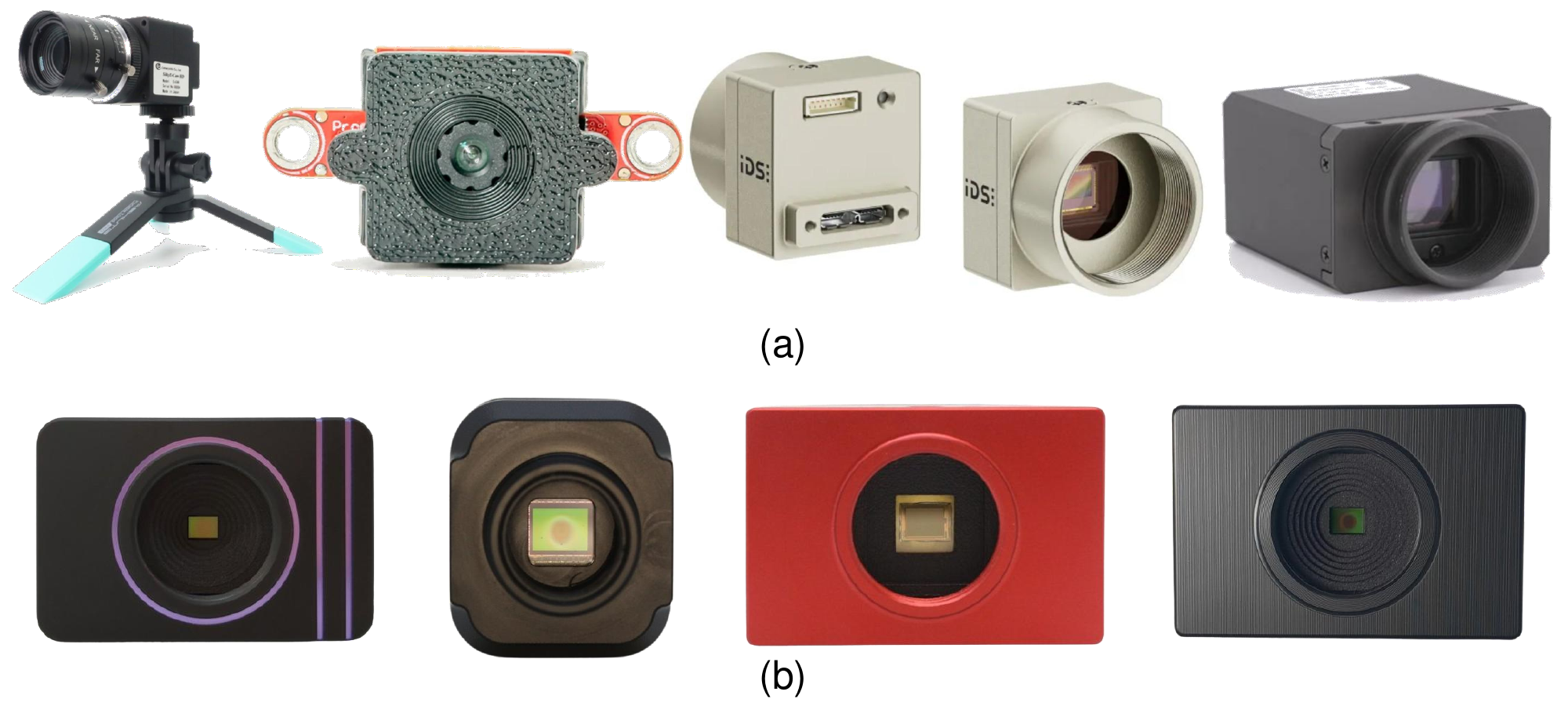}
\caption{Off-the-shelf event cameras. (a) Prophesee family of event cameras~\cite{prophesee_evaluation_kits}, (left to right): Century Arks SilkyEVCam, OpenMV GENX320 Camera Module,  IDS uEye XCP, and Lucid Triton2 EVS. (b) IniVation family of event cameras~\cite{inivation_buy}, (left to right): DVXplorer, DVXplorer Micro, DAVIS346, and DVXplorer Lite.}
\label{fig02}
\end{figure}

Commercially available event cameras, such as Prophesee's Metavision sensors~\cite{prophesee_evaluation_kits} (See Fig. \ref{fig02}(a)) and the iniVation DAVIS series~\cite{inivation_buy} (See Fig. \ref{fig02}(b)), have made the technology more accessible. However, the research community is still in its early stages of adopting event cameras across application domains. A key challenge remains the limited availability of annotated real-world event datasets, which hinders the development and benchmarking of robust event-based perception models.

\subsection{Event Camera Simulators}

The lack of large-scale, annotated real-world event datasets poses a significant challenge in developing robust event-based vision models. To overcome this, various event camera simulators have been developed to generate synthetic event streams from either rendered $3D$ scenes or conventional video inputs. These simulators serve as valuable tools for prototyping and benchmarking event-based algorithms across domains such as object detection, tracking, SLAM, and neuromorphic benchmarking.

Event camera simulators differ in their underlying methodologies, input data types, and levels of realism. Some simulators, like ESIM~\cite{rebecq2018esim}, employ adaptive sampling over rendered $3D$ scenes, producing accurate event timing based on photometric changes and including inertial and ground-truth data. Others, such as v2e~\cite{hu2021v2e}, synthesize events from frame-based video using a learned DVS pixel model. Blender-based tools like the DAVIS Simulator~\cite{mueggler2017event} and V2CE~\cite{zhang2024v2ce} simulate camera trajectories over $3D$ environments, providing dense annotations such as depth and camera calibration.

Recent simulators like DVS-Voltmeter~\cite{lin2022dvs} and PECS~\cite{han2024physical} incorporate physical and circuit-level noise models, improving realism for downstream learning tasks. Similarly, CARLA's built-in DVS~\cite{dosovitskiy2017carla} module (detailed in Section~\ref{CARLADVS}) enables event generation in photorealistic, interactive driving environments, uniquely supporting traffic-focused studies. A summary of key simulators is presented in Table~\ref{table01}.
These tools differ in terms of fidelity, accessibility, and flexibility. As the field advances, a key research direction involves bridging the realism gap between synthetic and real-world events, making simulators increasingly important for pretraining and evaluation in low-data settings.

\begin{table}[t]
\centering
\caption{Overview of Event Camera Simulators}
\label{table01}
\resizebox{\columnwidth}{!}{%
\begin{tabular}{c|c|c|c|c|c}
\toprule
\textbf{Year} & \textbf{Simulator} & \textbf{Open-source} & \textbf{Language} & \textbf{Input} & \textbf{Output} \\
\midrule
2024 & V2CE~\cite{zhang2024v2ce} & \cmark & Python & RGB/grayscale video & Event streams \\
\hline
2024 & PECS~\cite{han2024physical} & \xmark & Python & Blender scenes & Photocurrent events (ray-traced) \\
\hline
2024 & ADV2E~\cite{jiang2024adv2e} & \xmark & Python & Synthetic circuit models & DVS circuit behavior events \\
\hline
2023 & Prophesee VtoE~\cite{prophesee_viz_video2event} & \xmark & Python & Images or video (png/jpg, mp4) & Event-based video/frame \\
\hline
2022 & DVS-Voltmeter~\cite{lin2022dvs} & \cmark & Python & High frame-rate video & Event streams with noise modeling \\
\hline
2021 & v2e~\cite{hu2021v2e} & \cmark & Python & RGB/grayscale video & Events (.h5 format) \\
\hline
2021 & ICNS Simulator~\cite{joubert2021event} & \cmark & Python/C++/MATLAB & Blender scenes, camera trajectory & Events, images, ground truth \\
\hline
2021 & EventGAN~\cite{zhu2021eventgan} & \cmark & Python & Training videos & GAN-generated event frames \\
\hline
2018 & ESIM~\cite{rebecq2018esim} & \cmark & C++ & 3D camera motion, initial images & Events, images, IMU, ground truth \\
\hline
2017 & DAVIS Simulator~\cite{mueggler2017event} & \cmark & Python & Camera trajectory, Blender scenes & Events, images, depth, calibration \\
\hline
2017 & CARLA DVS~\cite{dosovitskiy2017carla} & \cmark & Python/C++ & Rendered RGB frames in CARLA & carla.DVSEventArray (events) \\
\hline
2017 & PIX2NVS~\cite{bi2017pix2nvs} & \cmark & Python & Video frames & Event streams \\
\bottomrule

\end{tabular}%
}
\end{table}

\subsection{CARLA DVS}
\label{CARLADVS}
The CARLA simulator~\cite{dosovitskiy2017carla} provides native support for event-based vision through its \texttt{sensor.camera.dvs} blueprint, which emulates a dynamic vision sensor. Unlike conventional cameras, DVS asynchronously capture changes in brightness at each pixel, outputting a stream of events encoded as $e = \langle x, y, t, p \rangle$, where $(x, y)$ is the pixel location, $t$ is the timestamp, and $p$ is the polarity (+1 or -1).

Events are triggered when the change in logarithmic intensity at a pixel exceeds a predefined threshold:
\[
L(x, y, t) - L(x, y, t - \delta t) = p \cdot C
\]
where $C$ is the contrast threshold. The CARLA's DVS mimics this behavior by sampling differences between successive frames at high frequency, generating more events under fast motion or dynamic lighting. To approximate the microsecond temporal resolution of real event cameras, the DVS must run at higher frequencies than standard sensors, balancing temporal fidelity and computational cost.

\begin{figure}[t]
\centering
\includegraphics[width=0.90\linewidth]{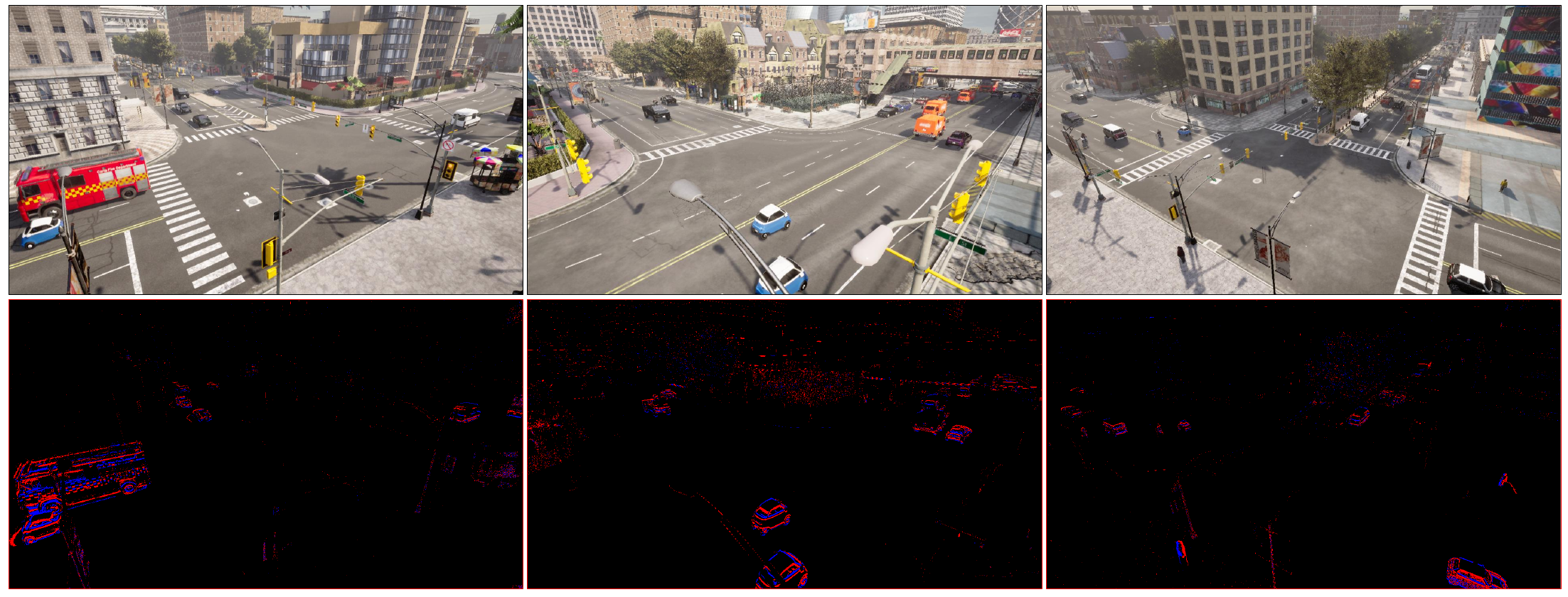}
\caption{Event frames (top row) generated by CARLA’s DVS and their corresponding RGB frames (bottom row), shown across different viewpoints with traffic objects.}
\label{fig03}
\end{figure}

The DVS camera in CARLA supports several configurable parameters, including contrast thresholds, refractory periods, and noise models. The event stream can be accessed using Python APIs and visualized as event frames, where positive and negative events are typically rendered in blue and red, respectively (See Fig. \ref{fig03}). While the sensor inherits all RGB camera attributes, it also includes unique DVS-specific settings such as \texttt{use\_log} and noise controls.

CARLA’s DVS enables scalable, controlled generation of event data across diverse traffic conditions. Its tight integration with a high-fidelity simulation environment makes it particularly valuable for evaluating event-based perception models and studying the sim-to-real transfer in traffic monitoring scenarios.

\section{The Synthetic Event-based Traffic Monitoring Dataset}

To systematically evaluate the sim-to-real gap in event-based traffic object detection, we curated a synthetic dataset using the DVS module in the CARLA simulator. This dataset, termed \textit{SeTraM} (\textit{\textbf{S}ynthetic \textbf{e}vent-based \textbf{Tra}ffic \textbf{M}onitoring}), captures dynamic traffic scenarios across multiple urban intersections from a fixed overhead perspective. To enable fair and meaningful comparisons, we selected the real-world eTraM dataset~\cite{verma2024etram}, which similarly records event-based traffic activity using a fixed perception setup under comparable environmental conditions. This section introduces \textit{SeTraM}, describing its data generation pipeline, statistics, and the alignment strategy adopted to ensure compatibility with eTraM for sim-to-real performance analysis.

\begin{figure}[t]
\centering
\includegraphics[width=0.80\linewidth]{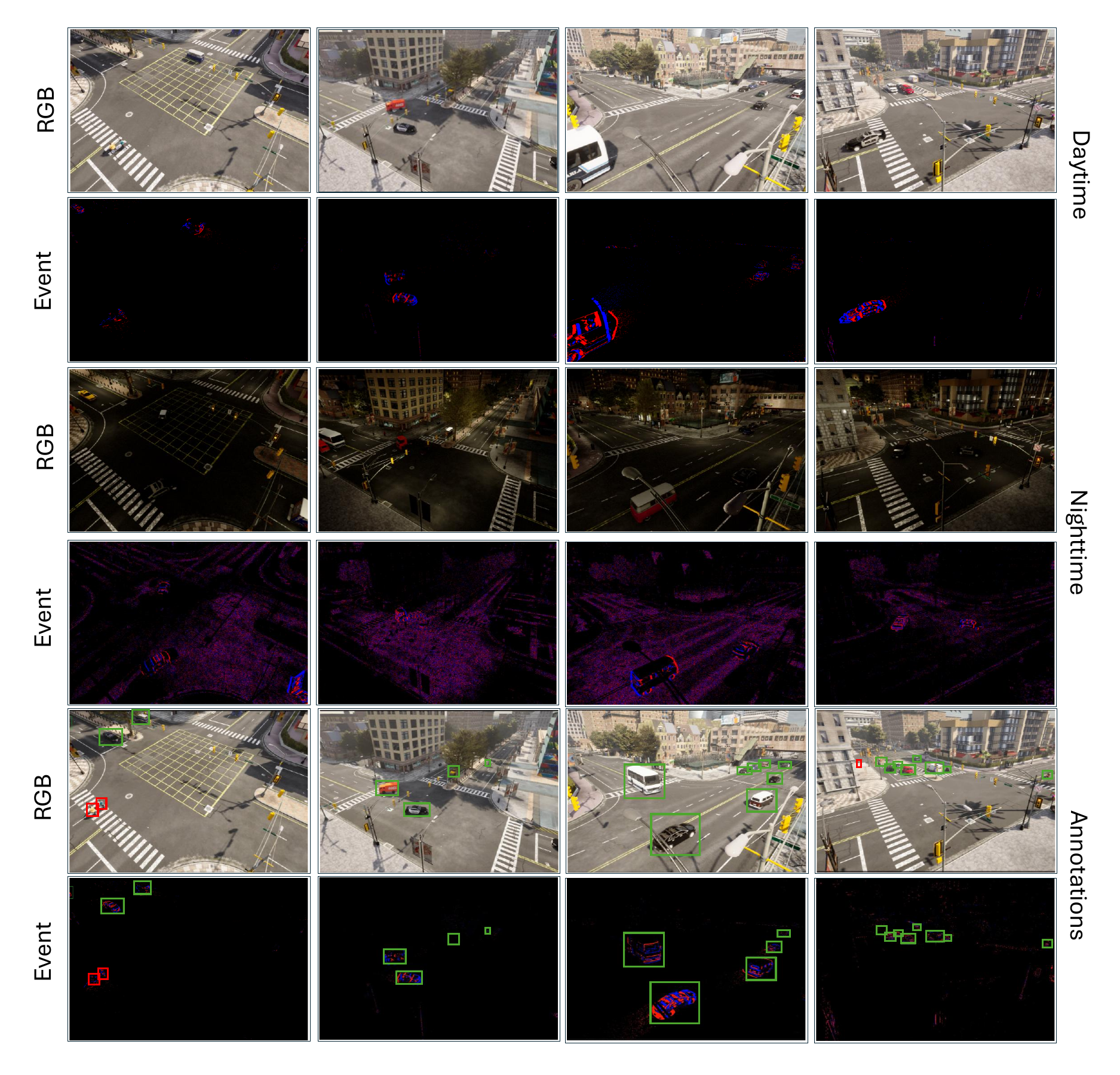}
\caption{Snapshots of \textit{SeTram} dataset showing daytime, nighttime and annotation instances.}
\label{datasetSamples}
\end{figure}

\subsection{\textit{SeTraM} Description}

The dataset used in this study is organized into seven groups, each comprising a fixed-length combination of synthetic and real-world data. The synthetic portion, derived from the \textit{SeTraM} dataset, was generated using the DVS module in the CARLA simulator and captures traffic activity from four distinct urban intersections, viewed from a fixed overhead perspective. Each group includes event sequences designed to simulate diverse real-world conditions by varying environmental factors such as lighting and traffic density (See Fig. \ref{datasetSamples}).

\textit{SeTraM} consists of five daytime and two nighttime sequences, each lasting approximately $80$ seconds, for a total of $38$ minutes of event data. To emulate realistic traffic dynamics, up to $100$ vehicles and $40$ pedestrians were deployed in each simulation. All sequences were annotated following the $1MPX$ format, with normalized bounding boxes and two object classes: pedestrian and vehicle.

The dataset was structured to support controlled sim-to-real experiments. All training sets maintain the same total duration but differ in the proportion of synthetic (\textit{SeTraM}) and real (eTram) data to assess model performance across varying domain mixes. Each training group comprises four sequences ($333$ seconds total). For validation, a consistent $50/50$ split between synthetic and real data was used. The test set includes two variants of eTram data: one entirely from daytime recordings, and another combining both day and night scenes, providing a robust basis for generalization evaluation.

\begin{figure}[t]
\centering
\includegraphics[width=0.90\linewidth]{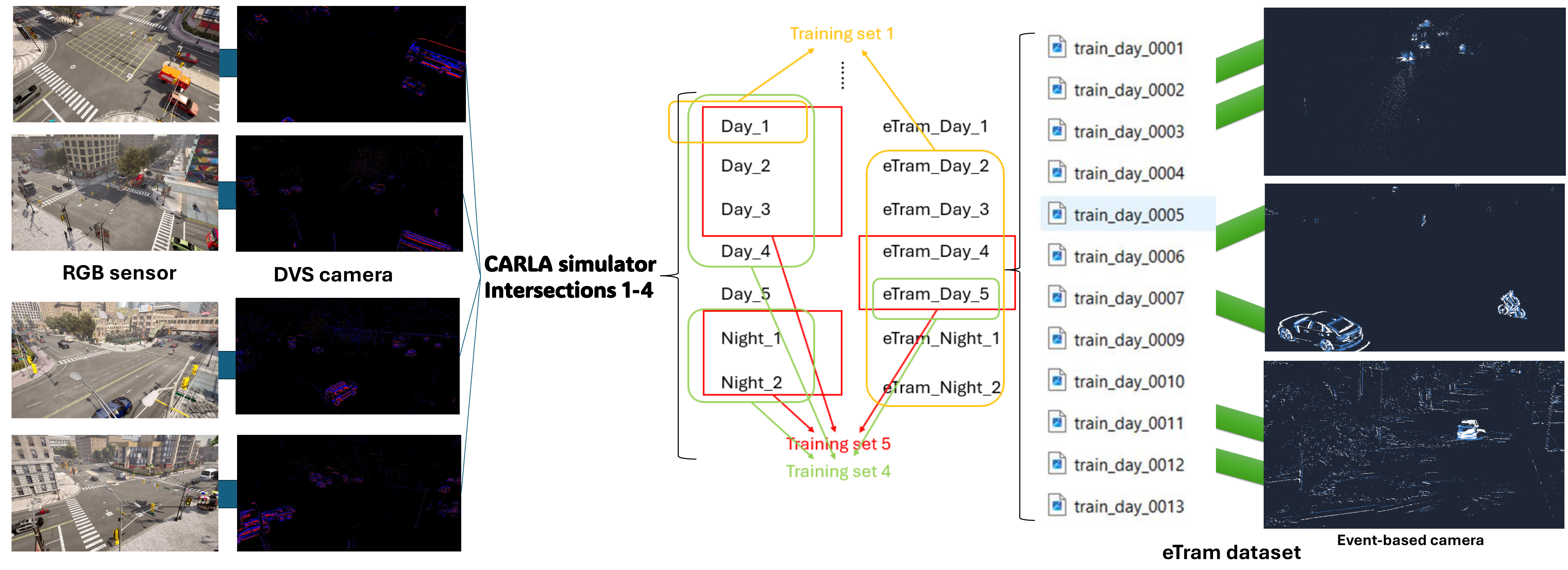}
\caption{Data generation pipeline for the training sets from the CARLA simulator at four-way intersections and the real-world eTram dataset}
\label{fig04}
\end{figure}

\subsection{Data Generation Pipeline}

The \textit{SeTraM} was generated through a structured pipeline designed to replicate realistic urban traffic environments. The overall procedure is summarized below (See Fig.~\ref{fig04}):

\begin{itemize} \item \textit{Traffic Simulation:} Using CARLA’s Python API, dynamic urban scenarios were constructed with moving vehicles and pedestrians. Four intersection layouts were selected, and traffic behavior was randomized across simulations to ensure diversity.

\item \textit{Sensor Deployment:} A fixed spectator was placed at the corner of each intersection, and both RGB and DVS sensors were attached to it with predefined transformations and orientations. Sensor parameters such as resolution (set to $1280$×$720$), polarity threshold, and day/night conditions were configured as needed.

\item \textit{Data Logging:} The simulator was executed in synchronous mode, advancing via \texttt{world.tick()} at each frame. Two queues were initialized to record RGB frames and DVS timestamps, which were later used for synchronization and annotation.

\item \textit{Bounding Box Processing:} During each simulation tick, bounding boxes were computed for actors (vehicles and pedestrians) located within a specified radius from the spectator. Only actors with velocity above 0.1 m/s were included in DVS annotations to ensure relevance, as stationary objects do not trigger events. Bounding boxes for RGB frames were stored in text files, and DVS bounding boxes were aligned using a timestamp synchronization mechanism. Specifically, CARLA’s DVS outputs events asynchronously in an array format via \texttt{listen()}, and all events within a frame's delta time interval were grouped and indexed using the starting timestamp.

\item \textit{Data Storage and Formatting:} DVS event streams were initially saved as \texttt{.csv} files and subsequently converted to \texttt{.npy} and \texttt{.h5} formats for compatibility with the RVT model. To ensure consistency with the eTram dataset and RVT pipeline, all annotations and event data were structured according to the 1MPX~\cite{perot2020learning} format. Bounding boxes followed a $30Hz$ appearance rate, with annotations consisting of seven fields: \textit{timestamp, x, y, w, h, class\_id}, and \textit{class\_confidence}~\cite{perot2020learning}.

\end{itemize}

The entire data generation process was controlled using a custom script, \texttt{generate\_traffic.py}, which allowed users to define simulation parameters such as duration, intersection layout, and number of actors. Upon completion, RGB frames and two CSV files (\texttt{output.csv} for events and \texttt{bbox.csv} for bounding boxes) were generated. These files were then processed using a custom conversion script, \texttt{rvtdataset.py}, to yield 1MPX-formatted \texttt{.npy} and \texttt{.h5} files. Finally, RVT’s preprocessing script was applied to convert these files into the format required for model training. The modularity of this pipeline allows for flexible adaptation to other data formats or perception tasks.

\subsection{Quantification of Data from Different Sources}

To ensure a fair comparison between synthetic and real-world data, all quantification in this study is based on temporal duration rather than the number of events or bounding boxes. Since the RVT processes event data within fixed timestamp intervals, aligning the datasets by time rather than volume avoids discrepancies caused by varying event densities across sources. This time-based standardization ensures consistency in model input and eliminates potential bias from data sparsity or density. The quantification methodology is described below (See Fig.~\ref{fig04} and Table~\ref{table02}):

\begin{table}[t]
\centering
\caption{Training Set Composition with Varying Proportions of eTram and \textit{SeTraM} Data}
\label{table02}
\begin{tabular}{c|c|c|c}
\toprule 
\textbf{Dataset ID} & \textbf{eTram (Real)} & \textbf{SeTraM (Synthetic)} & \textbf{Total Duration} \\
\midrule
Dataset \#1 & 0\% (0 sec)       & 100\% ($\sim$2300 sec)       & $\sim$2300 sec \\
Dataset \#2 & 15\% ($\sim$300 sec)    & 85\% ($\sim$2000 sec)        & $\sim$2300 sec \\
Dataset \#3 & 30\% ($\sim$600 sec)    & 70\% ($\sim$1600 sec)        & $\sim$2300 sec \\
Dataset \#4 & 45\% ($\sim$1000 sec)   & 55\% ($\sim$1300 sec)        & $\sim$2300 sec\\
Dataset \#5 & 55\% ($\sim$1300 sec)   & 45\% ($\sim$1000 sec)        & $\sim$2300 sec\\
Dataset \#6 & 70\% ($\sim$1600 sec)   & 30\% ($\sim$600 sec)         & $\sim$2300 sec \\
Dataset \#7 & 85\% ($\sim$2000 sec)   & 15\% ($\sim$300 sec)         & $\sim$2300 sec \\
\bottomrule
\end{tabular}
\end{table}

\begin{itemize} \item \textit{Data Instance Duration:} Each intersection instance in \textit{SeTraM} consists of $2500$ refresh cycles. With CARLA’s synchronous mode operating at a refresh interval of $0.0333$ seconds, each instance yields approximately $83$ seconds of event data.

\item \textit{Grouping of Instances:} A total of seven groups were constructed: five daytime groups and two nighttime groups. Each group comprises four instances (one from each of four intersections), resulting in a total length of $333$ seconds per group. These groups are labeled as Day 1–5 and Night 1–2.

\item \textit{eTram Data Alignment:} The original eTram recordings vary in duration, ranging from a few seconds to three minutes. To align with the \textit{SeTraM} structure, eTram sequences were re-segmented into seven groups of $S333$ seconds, maintaining the same distribution of five daytime and two nighttime groups. These are referred to as eTram-Day $1$–$5$ and eTram-Night $1$–$2$.

\item \textit{Construction of Mixed Datasets:} The initial dataset (Dataset \#1) was composed entirely of \textit{SeTraM} data. To study the impact of real data inclusion, eTram sequences were incrementally substituted into the training set in $1/7$ (14.3\%) steps. This resulted in seven training sets with increasing real-data proportions, ranging from 0\% to 85.7\%. The validation and test sets remained fixed across all experiments, with 52.9\% of the data coming from eTram ($360$ seconds for validation, $180$ seconds for testing) and 47.1\% from \textit{SeTraM} ($320$ seconds for validation, $160$ seconds for testing). This ensured that the only changing variable was the composition of the training data.
\end{itemize}

\begin{table}[t]
\centering
\begin{minipage}{0.48\linewidth}
\caption{Validation performance across models trained with varying proportions of real-world data.}
\label{table03}
\centering
\begin{tabular}{c|c|c|c|c}
   \toprule
    \textbf{ID} & \textbf{Real} & \textbf{AP@75} & \textbf{AP@50} & \textbf{mAP}\\ 
    \midrule
    1 & 0.000  & 20.21  & 24.61  & 18.28 \\  
    2 & 0.143  & 21.24  & 26.07  & 18.98 \\  
    3 & 0.286  & 20.24  & 67.77  & 30.53 \\  
    4 & 0.429  & 20.26  & 67.70  & 31.67 \\  
    5 & 0.571  & 30.27  & 55.54  & 31.23 \\  
    6 & 0.714  & 26.06  & 63.72  & 33.16 \\  
    7 & 0.857  & 4.81   & 31.24  & 12.21 \\ 
    \bottomrule
\end{tabular}
\end{minipage}
\hfill
\begin{minipage}{0.48\linewidth}
\caption{Test performance on mixed and night-only datasets using mAP and AP@50.}
\label{table04}
\centering
\begin{tabular}{c|c|c|c|c} 
  \toprule
    \textbf{ID} & \multicolumn{2}{c|}{\textbf{Mixed Test Set}} & \multicolumn{2}{c}{\textbf{Night-only Test Set}} \\ 
    ~ & \textbf{mAP} & \textbf{AP@50} & \textbf{mAP} & \textbf{AP@50} \\
    \midrule
    1 & 4.26 & 11.24 & 4.22 & 12.90 \\  
    2 & 6.63 & 16.46 & 5.94 & 16.43 \\  
    3 & 10.31 & 23.99 & 7.11 & 22.61 \\ 
    4 & 10.61 & 23.50 & 8.25 & 23.92 \\  
    5 & 12.18 & 28.01 & 12.19 & 30.52 \\  
    6 & 13.01 & 30.54 & 12.21 & 28.63 \\  
    7 & 15.69 & 36.09 & 12.63 & 30.81 \\  
    \bottomrule
\end{tabular}
\end{minipage}
\end{table}

\section{Experiments}
To isolate the effect of real-world data proportions on model performance, all the experimental settings were held constant across the training runs. The validation and test datasets remained identical across all experiments, ensuring that only the real/synthetic composition of the training data influenced the outcomes. Each model was trained and evaluated under the same computational environment and parameter configurations.

\subsection{Experimental Setup}

A total of seven models were trained using datasets \#1 through \#7 (refer to Table~\ref{table02}), each corresponding to a different ratio of real (eTram) and synthetic (\textit{SeTraM}) event data. All models were based on the RVT-Small architecture and trained using two $A100$ GPUs. The training configuration included a batch size of $5$ per GPU, $6$ data loading workers for training, and $2$ for evaluation. The learning rate was fixed at $0.0002236$, and training was conducted for a maximum of $400K$ steps.

\begin{figure}[t]
\centering
\includegraphics[width=0.80\linewidth]{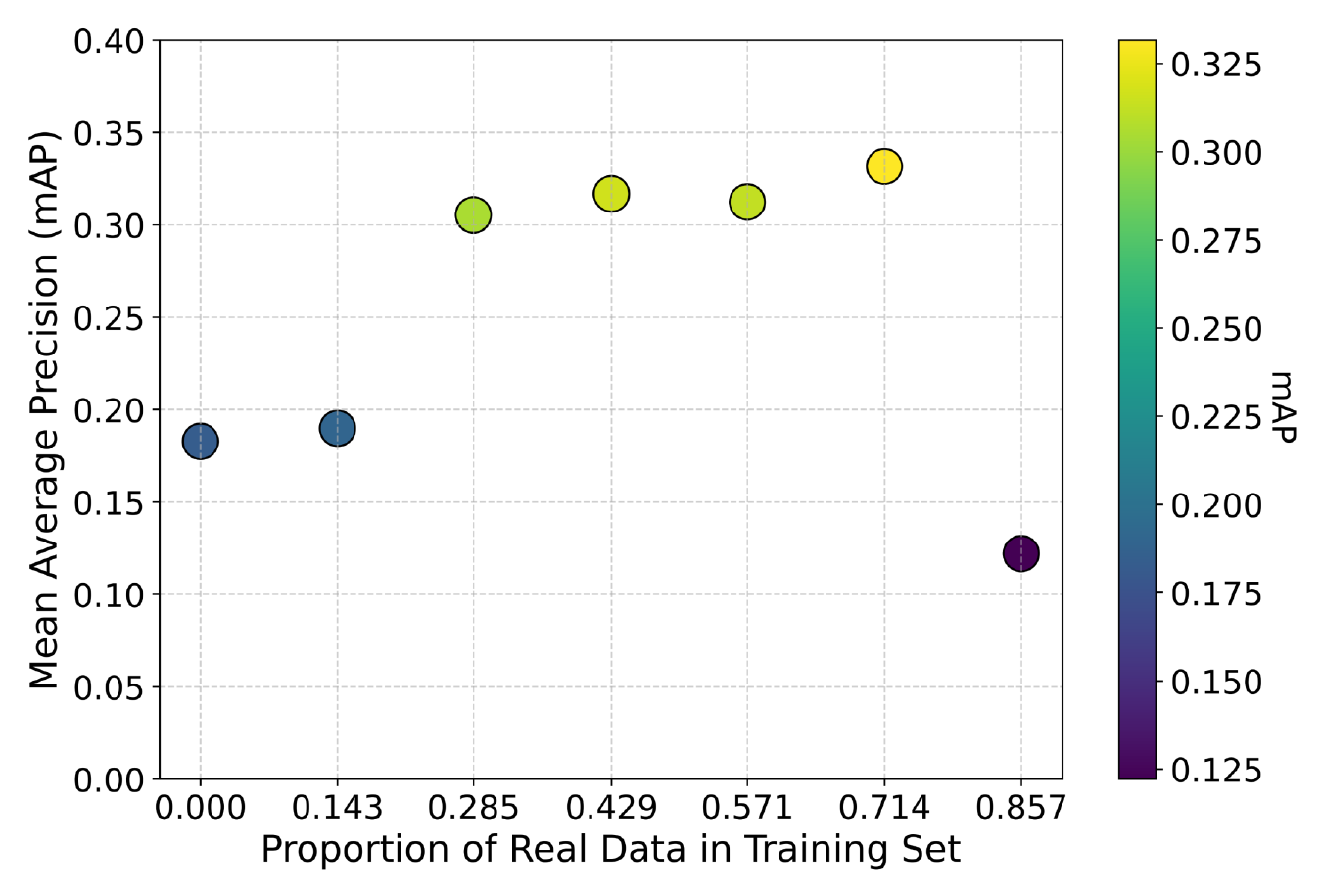}
\caption{Validation mAP across models with increasing real-world data proportions.}
\label{fig05}
\end{figure}

\subsection{Results and Analysis}

We evaluated model performance using three metrics: Average Precision at IoU thresholds of 50\% (AP@50) and 75\% (AP@75), and the overall mean Average Precision (mAP). As shown in Table~\ref{table03}, models trained with a balanced mix of synthetic and real data (e.g., Dataset \#3 to \#6) performed significantly better than models trained with either purely synthetic (Dataset \#1) or predominantly real (Dataset \#7) data. Interestingly, while the performance initially improved with increasing real data, it dropped sharply for Dataset \#7, suggesting overfitting or mismatch when the training data becomes heavily skewed toward real-world samples.

\begin{figure}[]
\centering
\includegraphics[width=0.90\linewidth]{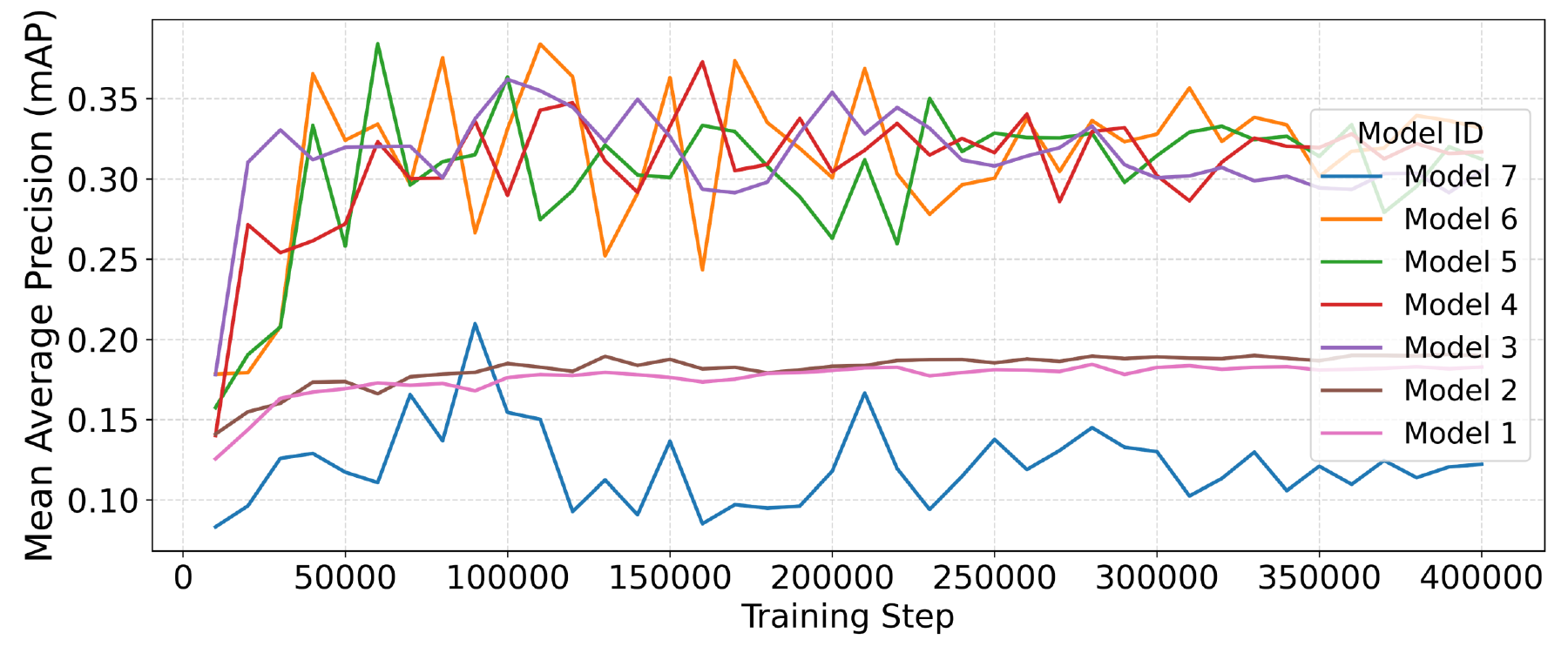}
\caption{Validation loss curves across different training configurations.}
\label{fig06}
\end{figure}

\begin{figure}[t]
\centering
\includegraphics[width=0.95\linewidth]{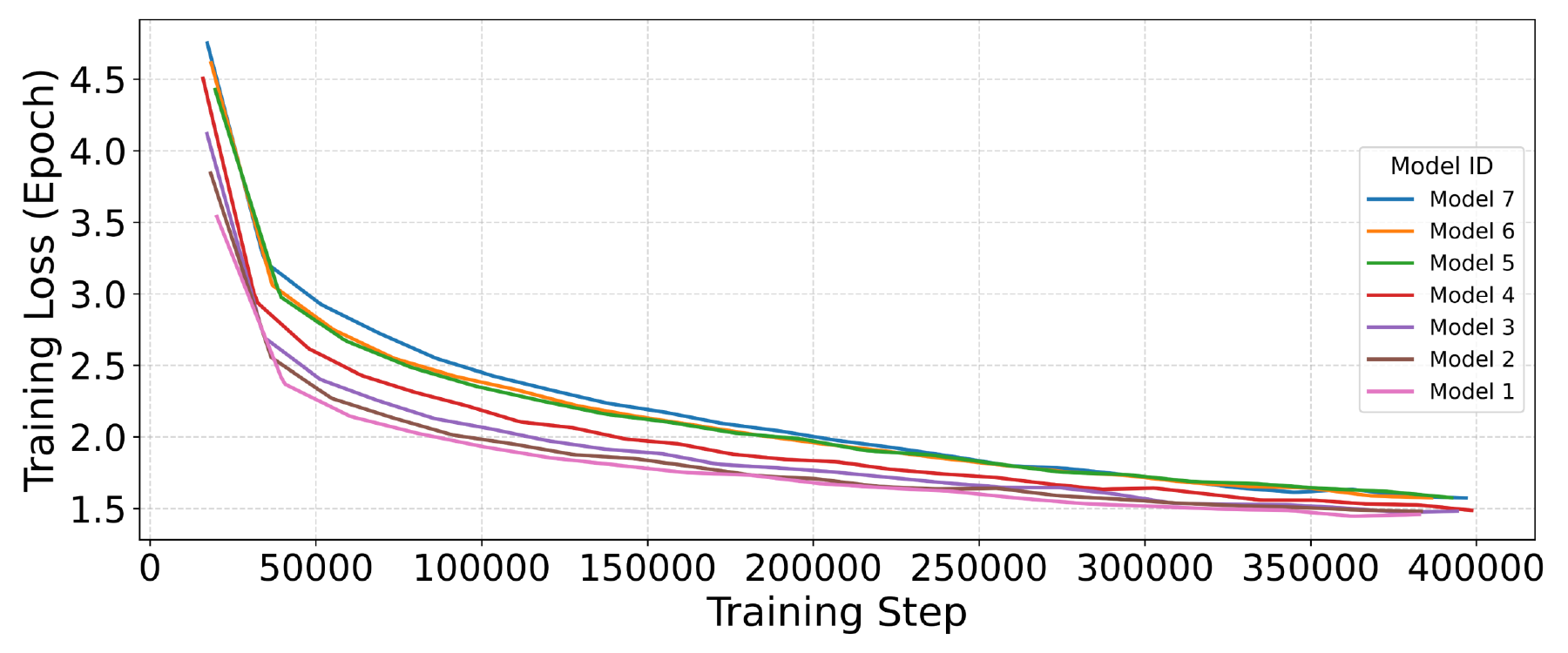}
\caption{Training loss vs. epochs for each dataset configuration.}
\label{fig07}
\end{figure}

\begin{figure}[]
\centering
\includegraphics[width=0.75\linewidth]{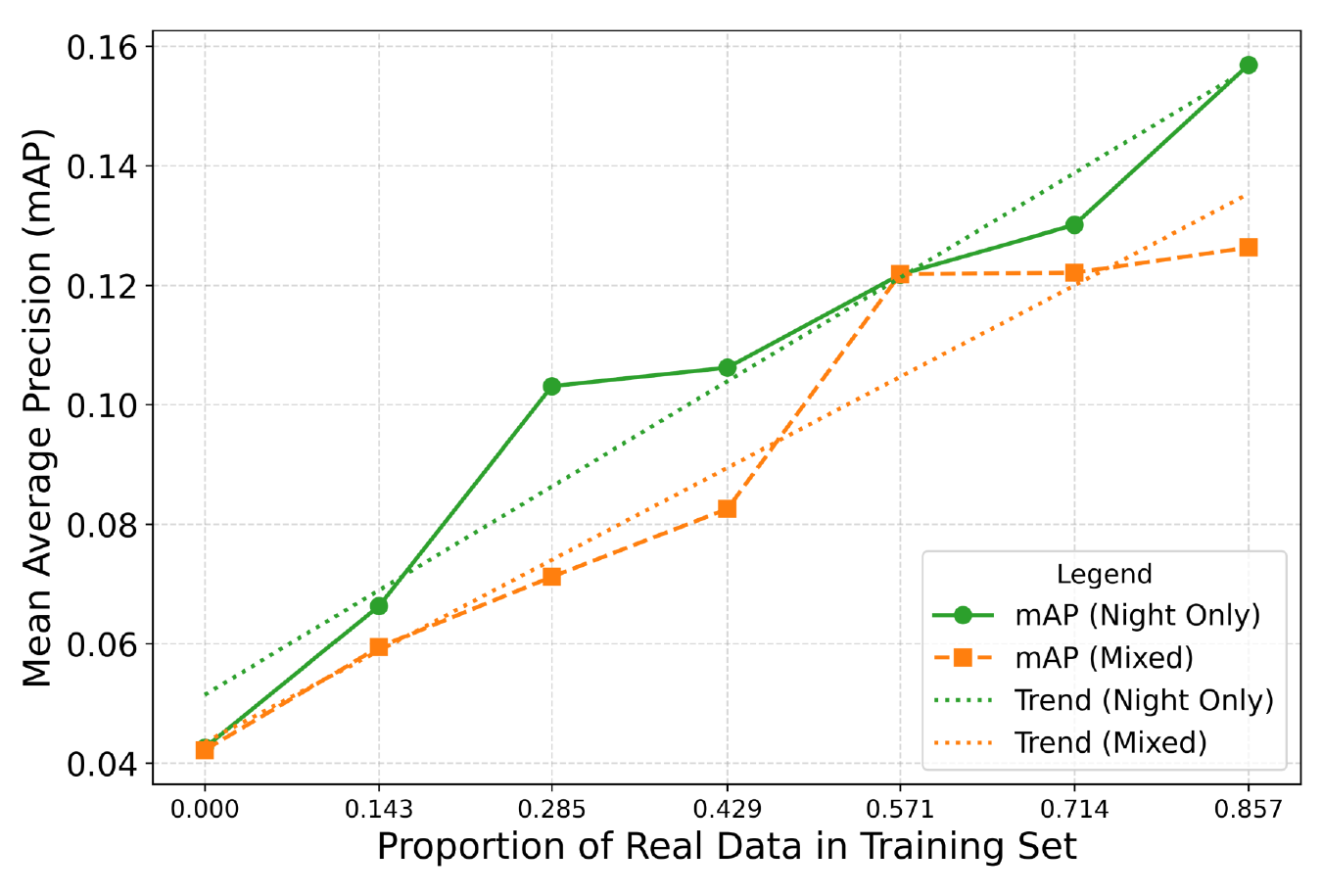}
\caption{Validation mAP across models with increasing real-world data proportions.}
\label{fig08}
\end{figure}

\begin{figure}[]
\centering
\includegraphics[width=0.81\linewidth]{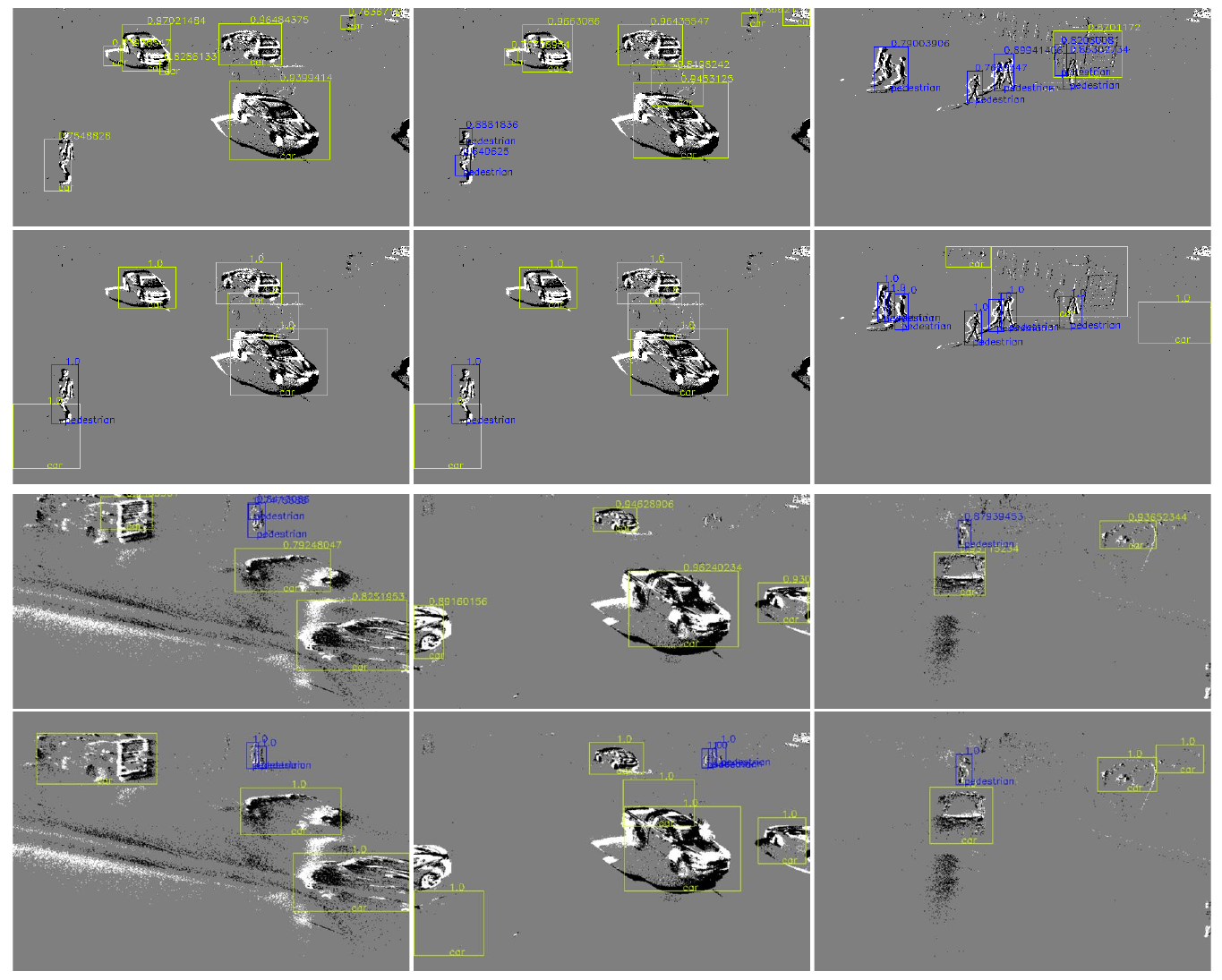}
\caption{Qualitative visualization of predictions for models $1$ to $7$. Ground-truth labels are shown in the bottom row, and model predictions are displayed in the top row.}
\label{fig09}
\end{figure}

The Fig.~\ref{fig05} shows the performance peak around Dataset \#5, after which the validation mAP declines. Fig.~\ref{fig06} and \ref{fig07} provide insights into model convergence and stability.

Training loss steadily decreased across all runs, indicating effective optimization, while validation loss fluctuated more substantially, likely due to domain shifts between synthetic and real distributions.
The validation set contains 52.9\% real and 47.1\% synthetic data. Models trained with a similar distribution (Datasets \#4 to \#6) achieved higher validation performance, suggesting better generalization when domain alignment is preserved. However, when evaluated on fully real test data, the relationship between performance and the fraction of real training data becomes more linear, as shown in Fig.~\ref{fig08}.

Table~\ref{table04} and Fig.~\ref{fig08} show that model performance increases linearly with more real data on both fully real test sets. Interestingly, performance on the mixed (day + night) test set is generally higher than the night-only test set, suggesting a domain mismatch or lack of nighttime diversity in training. This indicates that while increasing real-world data can improve generalization, domain-specific variability (e.g., lighting conditions) still poses a challenge for sim-to-real transfer. Fig.~\ref{fig09} shows the qualitative visualization of traffic object detections.

\section{Discussion}
The experimental results reveal a nearly linear improvement in model performance as the proportion of real-world data in the training set increases. This trend indicates that real data offers consistent value in training robust object detection models, reinforcing the importance of authentic event streams for neuromorphic vision.
Despite leveraging high-fidelity synthetic data from CARLA’s DVS, models trained solely on synthetic inputs perform poorly when evaluated on real data. 
Conversely, performance improves markedly when real-world data is introduced, even in small increments, highlighting a significant domain gap. This gap is further evidenced by the instability in validation loss, suggesting poor generalization due to mismatched data distributions.

The validation curve shows an average slope of $0.115$ mAP per unit increase in real data proportion. Although this implies that synthetic data can capture some task-relevant structure, it also reveals a ceiling that models trained entirely on synthetic data consistently fall short by a sizable margin. Even with state-of-the-art models like RVT-base, maximum performance remains limited, with synthetic-only training unable to exceed a modest baseline.
It’s important to note that this sim-to-real gap may vary across tasks and architectures. While object detection is highly sensitive to domain shifts, other tasks, such as trajectory prediction or scene segmentation, may tolerate synthetic data better. Additionally, models equipped with domain adaptation techniques may help close the gap more effectively.  Evaluating this gap quantitatively requires dedicated sim-to-real gap metrics, such as the event quality score \cite{chanda2025event}. While CARLA-generated synthetic data (\textit{SeTraM}) is useful for controlled experimentation, it cannot yet replace real-world data in performance-critical event-based object detection. Improving simulation realism and leveraging cross-domain learning remain essential for effective sim-to-real transfer.

\section{Conclusion}
This study demonstrates that synthetic event data from CARLA’s DVS module falls short as a substitute for real-world event camera data in object detection. Despite some performance gains, a measurable sim-to-real domain gap persists, limiting generalization and overall model accuracy.
Our findings emphasize the need for better simulation fidelity, enhanced noise modeling, and alignment with real sensor behavior. Although \textit{SeTraM} provides a valuable benchmark, real data remain indispensable for high-performance event-based perception.
Future work should explore advanced domain adaptation methods, diverse architectures, and differences in data characteristics, such as event sparsity, absence of realistic noise, and temporal sampling uniformity in synthetic streams, which diverge from the complex and asynchronous nature of real event cameras to further bridge the sim-to-real gap and improve the practicality of synthetic data in real-world neuromorphic vision systems.

%
%
%

\bibliographystyle{splncs04}
\bibliography{main}

\end{document}